%% file: acl_latex.tex
\documentclass[11pt]{article}

\usepackage[preprint]{acl}

\usepackage{times}
\usepackage{latexsym}
\usepackage{enumitem}

\usepackage[T1]{fontenc}

\usepackage[utf8]{inputenc}

\usepackage{microtype}

\usepackage{inconsolata}

\usepackage{graphicx}

\usepackage{amsmath}
\usepackage{xcolor}
\definecolor{linkblue}{HTML}{0000EE}

\usepackage{tcolorbox}
\tcbuselibrary{skins}
\definecolor{insightblue}{HTML}{0047AB}
\newtcolorbox{insightbox}{%
  enhanced,
  colback=blue!3,
  colframe=insightblue,
  boxrule=0.8pt,
  arc=1pt,
  left=4pt,
  right=4pt,
  top=1pt,
  bottom=1pt,
  before skip=6pt,
  after skip=4pt,
}

\makeatletter
\AtBeginDocument{\setlength{\@fptop}{0pt}\setlength{\@dblfptop}{0pt}}
\makeatother

\title{Demystifying Training-Time Augmentation for Data-Constrained Language Model Pretraining}

\author{
  {\bf Michael K. Chen}$^1$ \ {\bf Xikun Zhang}$^2$ \\
  {\bf Fan Bai}$^3$ \ {\bf Zhengding Hu}$^1$ \ {\bf Zhen Wang}$^1$ \\
  $^1$UC San Diego $^2$RMIT University $^3$Johns Hopkins University \\
  \small \texttt{\{mkc013, zhw085\}@ucsd.edu}
}

\begin{document}
\maketitle
\begin{abstract}
As AI labs approach a data ceiling where compute capacity outpaces the rate of new high-quality text generation, language model pretraining is shifting toward a data-constrained, compute-abundant regime that demands productive multi-epoch training on fixed corpora. Standard autoregressive (AR) pretraining overfits severely in this setting, reaching its optimum early and then continuously deteriorating. We investigate training-time data augmentation as a regularizer to mitigate this overfitting and enable productive training for hundreds of epochs on the same data. We introduce three orthogonal categories of augmentation for AR pretraining: token-level noise (masking, random replacement), sequence permutations (right-to-left prediction, Fill-in-the-Middle), and target offset prediction ($x_{t+i}$ for $i > 1$). Through systematic ablations, we find that individual augmentations delay overfitting and lower validation loss relative to the baseline, with random token replacement achieving the best minimum loss among individual methods. Combining augmentation categories further lowers the minimum validation loss. Our experiments demonstrate that data augmentations mitigate AR pretraining's data inefficiency and offer a promising solution to the data-constrained regime~\footnote{All code and data are available at \href{https://github.com/michaelchen-lab/data-augmentations-for-pretraining}{\color{linkblue}{https://github.com/ michaelchen-lab/ data-augmentations-for-pretraining}}}.
\end{abstract}

\section{Introduction}
\label{sec:intro}

For a decade, language model (LM) pretraining has improved by scaling model, data, and compute together \cite{kaplan2020scaling, hoffmann2022chinchilla}. That recipe is now supply-limited: compute keeps growing, while the stock of high-quality human text is projected to be largely exhausted within a few years \cite{villalobos2022will, muennighoff2023scaling}. Pretraining is moving into a \textit{compute-abundant, data-constrained} regime, where the binding constraint shifts from how many tokens a model can process to how much generalizable signal it can extract from each one. A fixed corpus must then be revisited for many epochs, making repeated-data training a central problem.

Many-epoch training over a fixed corpus exposes a failure mode of autoregressive (AR) next-token prediction: trained long enough on the same tokens, the model shifts from generalizing to memorizing, and held-out loss rises \cite{hernandez2022scaling, tirumala2022memorization, xue2023repeat}. The damage exceeds diminishing returns: past a few epochs the value of repeated data falls toward zero \cite{muennighoff2023scaling}, and deeper into the regime continued training drives held-out loss back up, undoing earlier progress \cite{kim2025pre}. Diffusion offers a contrast: a diffusion model learns to reconstruct data from many sampled noise levels, so across training each sequence is seen through a continually changing set of corrupted views. Recent diffusion language models, which carry this objective to text, resist repeated-data overfitting markedly better than AR models in the same regime \cite{prabhudesai2026diffusion, ni2025diffusion}. However, replacing AR with a diffusion language model wholesale is impractical today: it abandons the autoregressive next-token formulation that the whole training and inference stack is built around, and brings well-documented costs, including many sampling steps for competitive quality \cite{feng2026theoretical}, awkward variable-length generation \cite{li2025survey}, and immature infrastructure \cite{peng2025efficient}. The robustness itself, though, is usually attributed to a concrete property of the objective: each sequence is presented as many varied, stochastic views, so the model rarely solves the same prediction problem twice. This property can be approximated inside a standard AR model by augmenting its training data, with no change to the architecture.

\begin{figure*}[t]
  \centering
  \includegraphics[width=\textwidth]{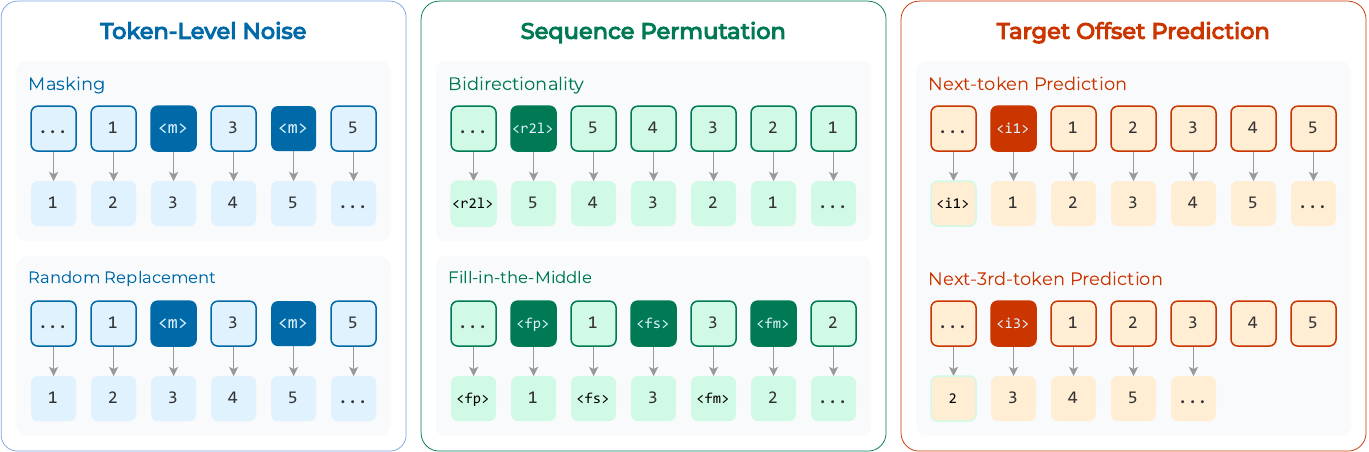}
  \caption{Overview of the three augmentation categories. Each panel shows an example (input, label) pair under the named transformation. \textbf{Token-level noise} (left) replaces a fraction of input tokens with a mask token or a random vocabulary token; labels are always the original uncorrupted sequence. \textbf{Sequence permutations} (middle) reverse the sequence for right-to-left prediction (R2L) or reorder it as prefix--suffix--middle suffix--prefix--middle for Fill-in-the-Middle (FIM); labels match the rearranged order. \textbf{Target offset prediction} (right) trains the model to predict a future token $x_{t+i}$ rather than just immediate next token; a prepended offset token indicates active horizon.}
  \label{fig:augmentations}
  \vspace{-10pt}
\end{figure*}

Individually, several such operations have been proposed before, token corruption \cite{devlin2019bert}, permuted factorization orders \cite{yang2019xlnet}, mixed denoising objectives \cite{tay2022ul2}, and infilling reorderings \cite{bavarian2022fim, nguyen2023meet}, but each in isolation and largely for single-epoch, compute-rich training. None connects them as instances of diffusion-style view generation, and none asks whether they regularize many-epoch, fixed-corpus AR pretraining, the regime the data wall now forces. Instead of reproducing diffusion training inside an AR model, we aim to identify which diffusion-style augmentations actually reduce overfitting when a standard AR model is trained for many epochs on a fixed corpus. We study three families, each varying one property of the predictive view (Figure~\ref{fig:augmentations}): \emph{input corruption} (token-level masking or random replacement, the AR analogue of denoising), \emph{factorization diversity} (right-to-left prediction and Fill-in-the-Middle, which relax the strict left-to-right order), and \emph{target diversity} (predicting a token at a sampled future offset instead of always the immediate next). Every augmentation applies only at training time and only to the (input, label) pair, leaving the architecture, the loss, and the left-to-right next-token evaluation unchanged; this training-time-only design is what separates the study from diffusion language models, which change the model itself and its inference procedure.

For experiment efficiency and scalability, we study a small 150M-parameter Llama-based model trained on 75M tokens of filtered web text, roughly $40\times$ below the Chinchilla-optimal budget \cite{hoffmann2022chinchilla, li2024datacomp}; behavior at larger model and data scales is left to future work. Held-out validation loss over many epochs is the primary metric, with zero-shot benchmarks \cite{eval-harness} as corroborating evidence. Four findings emerge.

\begin{itemize}[leftmargin=*, itemsep=0pt, topsep=0pt]
\item \textbf{Standard AR repetition fails early, then actively hurts.} The baseline reaches its lowest held-out loss around epoch~16 and degrades monotonically afterward, so most of a long run is counterproductive.
\item \textbf{Augmented views help only when they stay close to the left-to-right evaluation task.} Random token replacement beats masking and right-to-left prediction regularizes well, whereas Fill-in-the-Middle, whose format departs furthest from evaluation, gives no benefit.
\item \textbf{Predictive-target diversity helps only when anchored near next-token prediction.} A uniform distribution over a wide horizon erases useful signal, while an exponentially-weighted horizon that keeps most mass on the next token acts as an implicit curriculum.
\item \textbf{Composing augmentations yields the largest gains, and their interactions decide the outcome.} Token noise and offset prediction interfere when noise corrupts the local context offset prediction relies on, while right-to-left and offset prediction reinforce each other; the strongest configuration (low-rate random replacement~+~R2L~+~offset) lowers the minimum validation loss from 4.015 to 3.805, below every individual method and every naive stack.
\end{itemize}

\noindent Together, these results map which diffusion-style mechanisms transfer to AR, which fail, and which interfere, and they point to a practical direction for data-constrained pretraining: designing training-time predictive views that let a standard AR model keep improving over many epochs on a fixed corpus, extracting more from limited data.

\section{Method}
\label{sec:method}

We introduce three orthogonal categories of training-time data augmentation for autoregressive pretraining (Figure~\ref{fig:augmentations}). All augmentations modify the \emph{(input, label)} pair presented to the model at each training step; the architecture and loss function are unchanged. At evaluation time, all augmentations are disabled and the model is evaluated under the standard left-to-right (L2R) next-token prediction setting with $i = 1$. The three categories can be applied simultaneously; their composition is described in Section~\ref{sec:combining}.

\subsection{Category 1: Token-Level Noise}

Given a training sequence $(x_1, \ldots, x_L)$, a single uniform draw $u_t \sim \mathrm{Uniform}(0,1)$ per content token determines whether it is corrupted:
\begin{equation}
\tilde{x}_t = \begin{cases}
\texttt{<mask>} & \text{if } u_t < \alpha_m \\
x_{\mathrm{rand}} & \text{if } \alpha_m \le u_t < \alpha_m + \alpha_r \\
x_t & \text{otherwise,}
\end{cases}
\label{eq:noise}
\end{equation}
where $\alpha_m \in [0,1]$ is the mask rate, $\alpha_r \in [0,1]$ the random-replacement rate, and $x_{\mathrm{rand}}$ is sampled uniformly from the non-special vocabulary. The \emph{label} sequence is never modified: the model always predicts the original $x_t$, not $\tilde{x}_t$. Control tokens (direction, FIM, and offset tokens) are protected and never corrupted.

\noindent \textbf{Masking} ($\alpha_m > 0$, $\alpha_r = 0$): each selected token is replaced by a dedicated \texttt{<mask>} special token that never appears in unlabeled text. The model receives no lexical signal from masked positions and must recover original token from context alone.

\noindent \textbf{Random replacement} ($\alpha_m = 0$, $\alpha_r > 0$): each selected token is replaced by a random vocabulary token. Unlike masking, the replacement is a plausible but incorrect token, providing a semantically harder, yet more realistic signal.

\subsection{Category 2: Sequence Permutations}

\paragraph{Right-to-left (R2L) prediction.}
Each training sample is independently routed to L2R with probability $\rho$ or R2L with probability $1-\rho$. A direction token prepended at position 0 signals the mode:
\begin{align}
\text{L2R:} \quad & [\,\texttt{<l2r>},\; x_1,\; \ldots,\; x_{L-1}\,] \notag \\
\text{R2L:} \quad & [\,\texttt{<r2l>},\; x_L,\; x_{L-1},\; \ldots,\; x_2\,] \notag
\end{align}
In both cases, labels are the original or reversed token sequence, respectively, and the standard causal cross-entropy loss is applied. The direction token is prepended to L2R samples as well, so the input format is consistent.

\paragraph{Fill-in-the-Middle (FIM).}
Following \citet{bavarian2022fim}, each training sample is routed to PSM or SPM with probabilities $p_{\mathrm{psm}}$ and $p_{\mathrm{spm}}$ respectively, or left unchanged otherwise. Two pivot positions $a < b$ are sampled uniformly, splitting the content (the first $L_c = L-3$ tokens, reserving 3 positions for FIM control tokens) into:
\begin{align}
P &= (x_1, \ldots, x_a), \notag \\
M &= (x_{a+1}, \ldots, x_b), \notag \\
S &= (x_{b+1}, \ldots, x_{L_c}). \notag
\end{align}
The segments are rearranged using control tokens \texttt{<fp>}, \texttt{<fs>}, \texttt{<fm>}:
\begin{align}
\text{PSM:} \quad & [\,\texttt{<fp>},\; P,\; \texttt{<fs>},\; S,\; \texttt{<fm>},\; M\,] \notag \\
\text{SPM:} \quad & [\,\texttt{<fp>},\; S,\; \texttt{<fs>},\; P,\; \texttt{<fm>},\; M\,] \notag
\end{align}
Labels equal the rearranged sequence, so the model is trained to predict every token in the rearranged left-to-right order. When the model arrives at \texttt{<fm>}, it has already seen both prefix and suffix and must predict the middle, i.e., the fill-in-the-middle objective.

\subsection{Category 3: Target Offset Prediction}

Rather than predicting the immediately next token $x_{t+1}$, the model predicts $x_{t+i}$ where offset $i$ is sampled once per training sample from a distribution over $\{1, \ldots, n\}$. Two weighting schemes were studied:
\begin{align}
P_{\mathrm{unif}}(i) &= \tfrac{1}{n}, \notag \\
P_{\mathrm{exp}}(i) &\propto e^{-(i-1)/T},
\label{eq:weighting}
\end{align}
with temperature $T = 1$. The exponential scheme concentrates mass on small offsets (especially $i=1$) while still sampling larger horizons, gradually extending the prediction range as a form of implicit curriculum.

A per-sample offset token \texttt{<next\_i>} is prepended, giving the layout:
\[
[\,\texttt{<next\_i>},\;x_1,\;\ldots,\;x_{L-1}\,],
\]
and the label at position $t$ is $x_{t+i}$ (positions where $t{+}i > L$ are masked with $-100$). When combined with R2L augmentation (Section~\ref{sec:combining}), a direction token is additionally prepended before \texttt{<next\_i>}, consuming one extra sequence position. At evaluation time the offset is fixed to $i=1$, restoring standard next-token prediction.

\subsection{Combining Augmentations}
\label{sec:combining}

The three categories compose as a sequential pipeline, applied in the following order at each training step:
\begin{enumerate}[leftmargin=*, itemsep=0pt, topsep=0pt]
\item \textbf{Token noise} corrupts content tokens in the raw input (Eq.~\ref{eq:noise}).
\item \textbf{FIM} (if active) rearranges the potentially noisy sequence into PSM or SPM. FIM control tokens are added \emph{after} token noise, so they are not subject to corruption.
\item \textbf{Direction / offset}: the direction token and offset token \texttt{<next\_i>} are prepended, the sequence is reversed if R2L, and labels are constructed.
\end{enumerate}
On any given step a sequence is routed to exactly one variant in stage 2 (L2R, R2L, PSM, or SPM), while stages 1 and 3 apply independently. Category 2 and Category 3 share the direction token slot: FIM samples always use L2R direction (no reversal) since the rearrangement is already non-trivial.

\smallskip\noindent\textit{Example} (random15 + R2L + $i{=}2$, source $(x_1,\ldots,x_6)$):
\begin{quote}
\textbf{After noise:} $(x_1,\hat{x}_2,x_3,x_4,x_5,x_6)$ {\small($\hat{x}_2\!\ne\!x_2$)} \\
\textbf{After R2L + offsets:} \texttt{<r2l>} \texttt{<next\_2>} $x_6\;x_5\;x_4\;x_3$ \\
\textbf{Labels:} $\text{-}100\;\text{-}100\;x_4\;x_3\;x_2\;x_1$
\end{quote}
At evaluation, token noise is off, no FIM, direction is L2R, and $i=1$: plain autoregressive decoding.

\section{Experimental Setup}
\label{sec:setup}

\textbf{Model.} We train a 150M-parameter causal language model based on the Llama architecture \cite{touvron2023llama}, implemented with the HuggingFace Transformers library \cite{wolf2020transformers}. The model has 20 transformer layers, a hidden size of 512, 4 attention heads, an intermediate size of 1536, and a maximum context length of 2048 tokens. Tied input/output embeddings are used to keep the parameter count tractable. We train with the Warmup-Stable-Decay (WSD) learning rate schedule \cite{hagele2024scaling}, which decouples a constant stable training phase from a short final decay. Following \citet{hagele2024scaling}, WSD achieves validation loss comparable to a cosine schedule while reducing training costs by allowing the stable-phase checkpoint to be reused across multiple decay restarts. We use a peak learning rate of $6 \times 10^{-4}$, 100 linear warmup steps, and weight decay of 0.033 with the AdamW optimizer. For ablation studies where the primary interest is relative comparison rather than absolute performance, we report validation loss directly from the stable phase without applying the final decay, enabling cheap and consistent comparisons across a larger number of runs. We validate our methodology with a robustness check in Section \ref{sec:decay}.

\noindent \textbf{Dataset.} We train on 75M tokens extracted from DCLM-RefinedWeb \cite{li2024datacomp}, a high-quality filtered web-text corpus. At the Chinchilla-optimal token-to-parameter ratio \cite{hoffmann2022chinchilla}, a 150M-parameter model would require approximately 3B training tokens; our 75M-token corpus therefore places us roughly $40\times$ below the compute-optimal budget, intentionally targeting the data-constrained regime. We use the Qwen2 tokenizer, extended with augmentation-specific special tokens if necessary: a direction pair (\texttt{<l2r>}/\texttt{<r2l>}), three FIM control tokens (\texttt{<fp>}/\texttt{<fs>}/\texttt{<fm>}), a mask token (\texttt{<mask>}), and per-offset tokens (\texttt{<next\_i>} for each $i \leq n$).

\noindent \textbf{Evaluation.} Our primary metric is held-out validation loss under the standard L2R next-token prediction objective ($i=1$), evaluated at regular checkpoints throughout training. This directly measures generalization quality and the onset and severity of overfitting. As a secondary metric, we evaluate zero-shot accuracy on five downstream benchmarks via \texttt{lm-evaluation-harness}: HellaSwag, PIQA, ARC-Challenge, WinoGrande, and COPA. At the 150M-parameter scale, zero-shot performance is noisy; we treat it as corroborating evidence rather than a primary signal.

\noindent \textbf{Training budget.} All ablation runs train for 100 epochs by default, which is sufficient to observe each method's minimum validation loss and the onset of overfitting. Runs are extended only when validation loss has not yet turned upward by epoch~100; since all runs eventually overfit monotonically, the minimum observed is the true minimum regardless of when training is stopped after that point, and the extended budget does not confer an advantage in minimum loss.

\section{Experiments}
\label{sec:experiments}

\subsection{Does standard pretraining overfit?}
\label{sec:baseline}

\begin{figure}[t]
  \centering
  \includegraphics[width=\columnwidth]{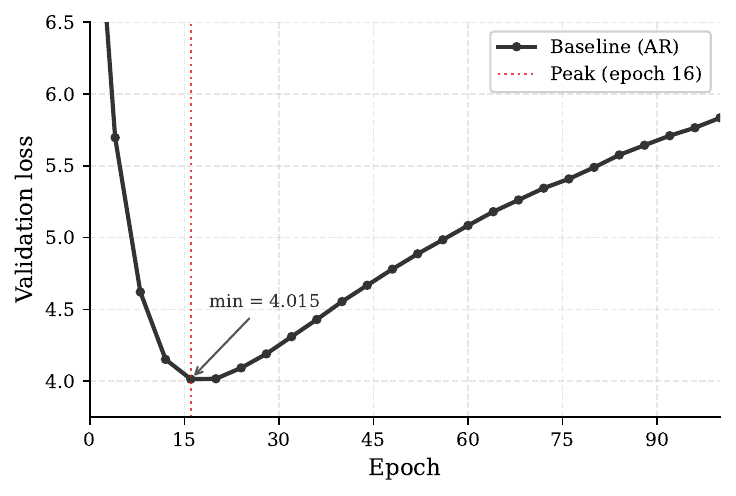}
    \vspace{-15pt}
  \caption{Validation loss of the baseline AR model over 100 epochs. The loss bottoms out at epoch~16 and deteriorates continuously thereafter.}
  \label{fig:baseline}
\vspace{-10pt}
\end{figure}

We begin by establishing the severity of overfitting under standard AR pretraining. Figure~\ref{fig:baseline} shows the validation loss trajectory of the baseline model over 100 epochs. The model reaches its minimum loss of 4.015 at epoch~16, after which loss increases monotonically for the remainder of training. Standard AR pretraining collapses into memorization within the first 20\% of training, rendering continued training counterproductive.

This trajectory is consistent with, and extends, the scaling laws of \citet{muennighoff2023scaling}, who find that repeated data yields negligible loss improvement up to about 4 epochs, with returns diminishing progressively toward zero for higher epoch counts. Our setting is substantially more extreme ($40\times$ below Chinchilla-optimal), and we observe not just diminishing positive returns but an \emph{actively harmful} phase: loss increases well above the single-epoch baseline, indicating that the model is driven toward memorization rather than generalization. This active degradation is not modeled by their scaling law, which describes the region of positive but diminishing returns. Our work can therefore be seen as probing the regime where that law breaks down, and augmentation becomes necessary. Our subsequent experiments test whether data augmentation can delay or prevent this collapse, while lowering validation loss.

\begin{insightbox}
\textbf{Finding 1.} In the data-constrained regime, naive repetition is self-defeating: standard AR pretraining stops improving early and then actively degrades, so the real question is how to keep more epochs useful rather than harmful.
\end{insightbox}

\subsection{Does token noise regularize, which kind?}
\label{sec:individual_token}

\begin{figure}[t]
  \centering
  \includegraphics[width=\columnwidth]{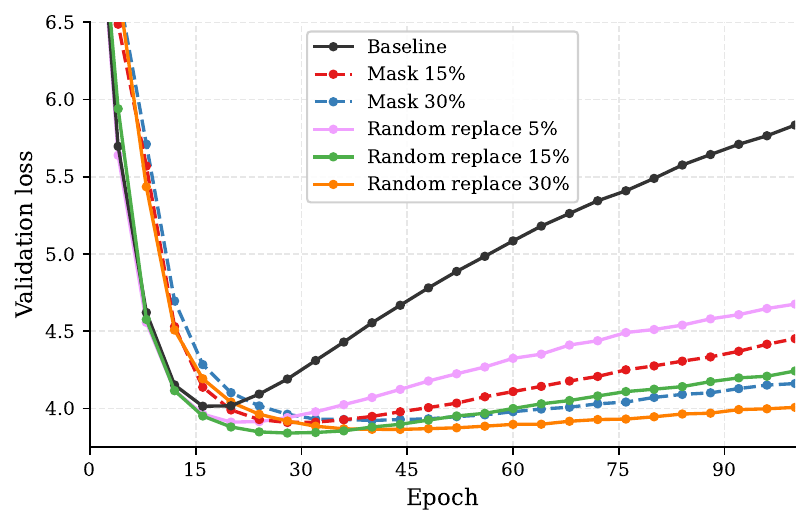}
      \vspace{-10pt}
  \caption{Validation loss for token-level noise ablations. Random replacement outperforms masking at matched rates; among random replacement variants, 15\% achieves the best individual minimum.}
  \label{fig:token_noise}
  \vspace{-10pt}
\end{figure}

We compare masking ($\alpha_m \in \{15\%, 30\%\}$) and random token replacement ($\alpha_r \in \{5\%, 15\%, 30\%\}$). Figure~\ref{fig:token_noise} shows all five variants against the baseline.
All four variants improve over the baseline (Table~\ref{tab:individual}). Random replacement consistently outperforms masking at matched corruption rates: random at 15\% achieves the lowest minimum loss among all individual methods (3.841 at epoch~28), compared to 3.910 for masking at 15\% at the same epoch. At the higher 30\% rate, both methods still improve over the baseline but with diminishing returns. A lower rate of 5\% also improves over the baseline (3.912 at epoch~20), but underperforms both higher-rate variants, indicating that 5\% noise is too weak a perturbation to provide effective standalone regularization. The advantage of random replacement is likely due to increased difficulty: a randomly replaced token is lexically plausible and must be identified as incorrect from context, whereas a masked token provides an explicit signal that information is absent. In Section~\ref{sec:combinations}, we conduct systematic combination experiments with random replacement at 15\% and 5\%.

\begin{insightbox}
\textbf{Finding 2.} The form of corruption matters: replacing tokens with plausible-but-wrong ones regularizes better than masking, because the model must judge correctness from context instead of detecting an obvious gap.
\end{insightbox}

\subsection{Which sequence permutations regularize?}
\label{sec:individual_perm}

\begin{figure}[t]
  \centering
  \includegraphics[width=\columnwidth]{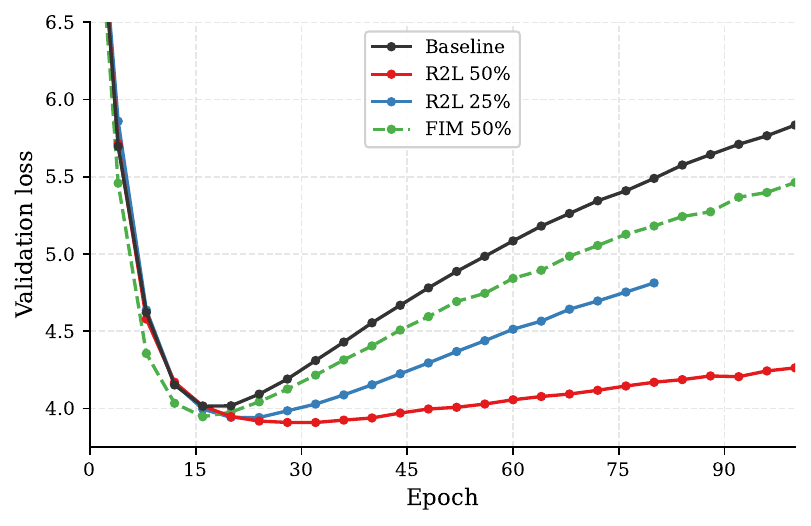}
    \vspace{-10pt}
  \caption{Validation loss for sequence permutation ablations. R2L at 50\% provides strong regularization; FIM provides essentially no benefit and overfits at the same rate as the baseline.}
  \label{fig:permutations}
  \vspace{-10pt}
\end{figure}

We compare R2L prediction at 25\% and 50\% mixing rates, and FIM at 50\% of samples (25\% PSM + 25\% SPM). Figure~\ref{fig:permutations} shows the results. R2L at 50\% outperforms R2L at 25\%, achieving a minimum loss of 3.910 at epoch~32 versus 3.942 at epoch~24. This suggests that a balanced direction split is more effective than a lopsided one: too little R2L (25\%) provides insufficient exposure to the reversed objective to regularize effectively.

FIM presents a clear negative result. Its minimum loss of 3.947 is reached at epoch~16, the same epoch the baseline bottoms out, and the loss then climbs steeply, surpassing the baseline by epoch~40. Despite FIM's utility as a code pretraining objective~\citep{bavarian2022fim}, it does not regularize general-domain text training effectively. We attribute this to a training--evaluation distribution mismatch: FIM rearranges sequences into formats so different from the standard L2R setting used at evaluation that the generalization benefit is limited. We adopt R2L 50\% as the preferred permutation variant and exclude FIM from combination experiments (Table~\ref{tab:individual}).

\begin{insightbox}
\textbf{Finding 3.} A useful training view stays close to the evaluation task: reversing a sequence still resembles next-token prediction and regularizes, whereas reformatting it for infilling diverges too far to help.
\end{insightbox}

\subsection{Does offset prediction help, and at what horizon?}
\label{sec:individual_offset}

\begin{figure}[t]
  \centering
  \includegraphics[width=\columnwidth]{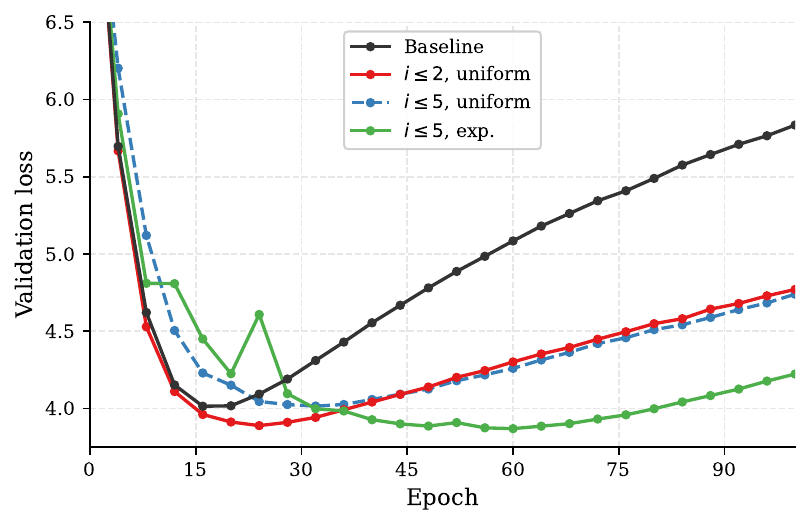}
    \vspace{-10pt}
  \caption{Validation loss for target offset prediction ablations. Exponential weighting over $i \leq 5$ is the strongest individual augmentation; uniform weighting over $i \leq 5$ provides no benefit.}
  \label{fig:offset}
    \vspace{-10pt}
\end{figure}

We compare uniform offset sampling over $\{1,\ldots,2\}$ and $\{1,\ldots,5\}$ with both uniform and exponential weighting. Results are shown in Figure~\ref{fig:offset}.
The weighting scheme is decisive. Uniform sampling over $i \leq 5$ produces essentially the same trajectory as the baseline (minimum loss 4.016 at epoch~32), with no regularization benefit despite the non-trivial offset targets. In contrast, exponential weighting over $i \leq 5$ is the strongest offset configuration: minimum loss 3.870 at epoch~60, with the loss still near its minimum at epoch~100. The key is that exponential weighting concentrates most probability mass on $i = 1$, so the model continues learning standard next-token prediction the majority of the time while occasionally training on harder longer-horizon targets. This implicit curriculum appears to be the effective regularizer; uniform sampling over a wide range degrades too much useful signal. The smaller $i \leq 2$ uniform variant (minimum 3.890 at epoch~24) confirms that a bounded offset alone can provide moderate regularization without exponential weighting. Notably, $i \leq 2$ reaches its best checkpoint 2.5$\times$ earlier than $i \leq 5$ exponential (epoch~24 vs.~60) with only a 0.020 gap in minimum loss, making it an attractive option when training budget is limited. We adopt $i \leq 5$ with exponential weighting as the preferred offset configuration, but revisit $i \leq 2$ in Section~\ref{sec:combinations}. However, as a practical caveat, we note that $i{\leq}5$ exp.\ is the only individual method that exhibits occasional spikes in the evaluation loss curve during the stable phase; it may therefore need closer monitoring than the other variants.

Table~\ref{tab:individual} collects the minimum validation loss and its corresponding epoch for all individual methods across the three augmentation categories.

\input{tables/tab_individual}

\begin{insightbox}
\textbf{Finding 4.} Diversity in the prediction target helps only when it stays anchored to next-token prediction: a curriculum that usually predicts the next token and occasionally reaches further regularizes, while spreading prediction uniformly across a wide horizon destroys the signal.
\end{insightbox}

\subsection{Do augmentations compose or interfere?}
\label{sec:combinations}

\begin{figure}[t]
  \centering
  \includegraphics[width=\columnwidth]{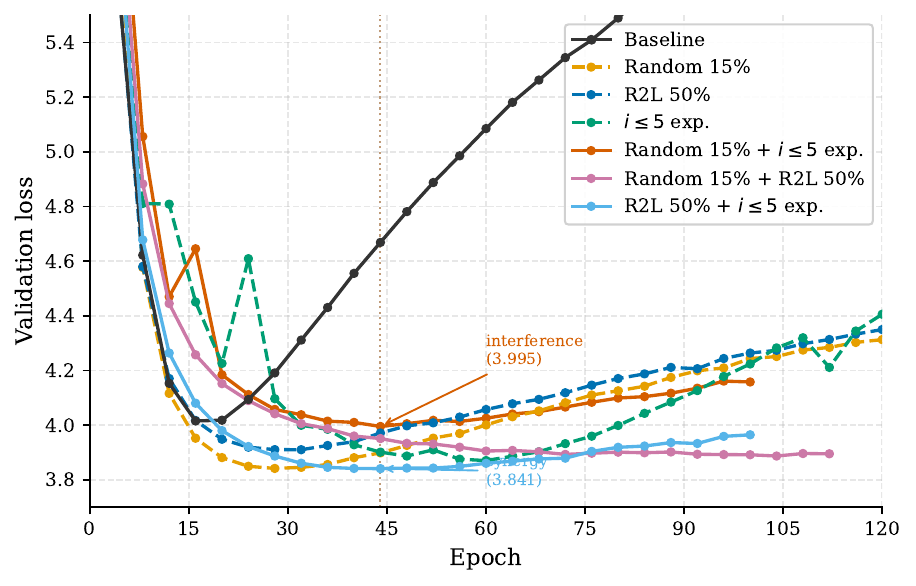}
    \vspace{-10pt}
  \caption{Systematic 2-cat combinations of the three best individuals. R2L~+~offset synergizes (sky blue, min 3.841). Token noise~+~offset interferes (red-orange, min 3.995). Token noise~+~R2L is intermediate (pink, min 3.887).}
  \label{fig:2cat}
    \vspace{-10pt}
\end{figure}

Table~\ref{tab:combinations} presents all 2- and 3-category combination results, organized by interaction type. A consistent pattern emerges across all runs: \emph{combinations involving token noise and offset prediction interfere strongly}, while \emph{R2L and offset prediction combine synergistically}. Token noise combined with R2L is intermediate, achieving a worse minimum loss than R2L alone but better than token noise combined with offset prediction. Figure~\ref{fig:2cat} shows the three systematic best-individual combinations.

\paragraph{Token noise $\times$ offset prediction: strong interference.}
All three token noise~$\times$~offset combinations fail badly. Random~15\%~+~$i{\leq}5$ exp.\ achieves a minimum of 3.995, nearly identical to the baseline (4.015). Mask~15\%~+~$i{\leq}2$ reaches only 4.004. Even at a third the noise rate, mask~5\%~+~$i{\leq}2$ achieves 3.937, still \emph{worse} than $i{\leq}2$ alone (3.890). The interference strengthens with higher noise rates, suggesting that high token corruption can meaningfully disrupt the offset objective. The mechanism is direct: offset prediction requires predicting $x_{t+i}$ from coherent local context at position $t$; token noise corrupts that context with plausible-but-wrong tokens, making the offset target near-unpredictable. The task becomes so hard that it degrades rather than regularizes optimization.

\paragraph{Permutation $\times$ offset prediction: synergy and an offset trade-off.}
R2L~50\%~+~$i{\leq}5$ exp.\ achieves a minimum of 3.841, tying the single best individual method. R2L preserves all token content and only reorders it, leaving the offset prediction task well-posed; the two objectives reinforce each other without interference. Replacing $i{\leq}5$ exponential with the cheaper $i{\leq}2$ uniform variant gives R2L~50\%~+~$i{\leq}2$: slightly worse minimum loss (3.863 vs.\ 3.841).

\begin{figure}[t]
  \centering
  \includegraphics[width=\columnwidth]{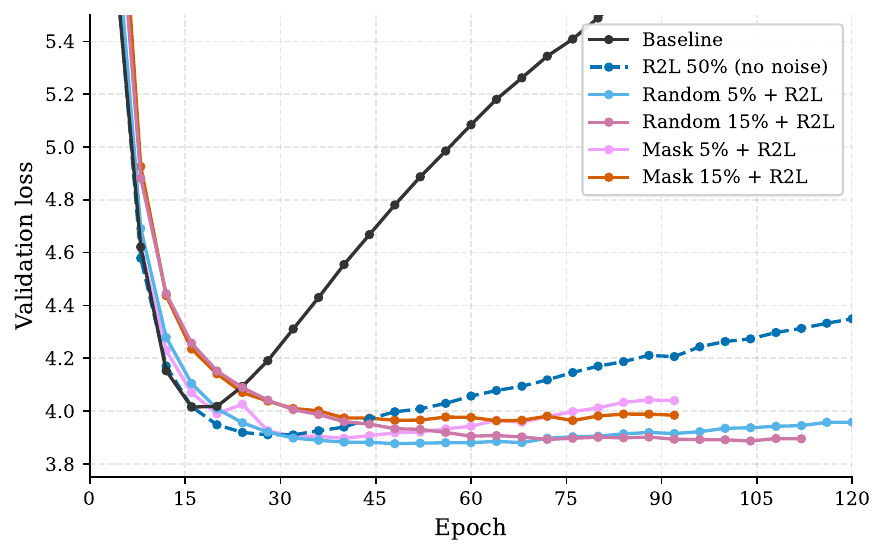}
    \vspace{-10pt}
  \caption{Token noise~$\times$~R2L combinations across noise rates and types. Random replacement consistently outperforms masking, and lower noise rates achieve better minimum loss while converging in fewer epochs.}
  \label{fig:noise_perm}
    \vspace{-5pt}
\end{figure}

\paragraph{Token noise $\times$ permutation: noise rate and type matter.}
Figure~\ref{fig:noise_perm} shows all four token noise~$\times$~R2L combinations. Random replacement consistently outperforms masking at the same rate: random~5\% (3.877) beats mask~5\% (3.897), and random~15\% (3.887) beats mask~15\% (3.963). Within each noise type, lower noise gives better minimum loss. Random~5\%~+~R2L achieves the best minimum of this group (3.877); random~15\%~+~R2L achieves a slightly higher minimum (3.887). Notably, mask~15\%~+~R2L~(3.963) is the only token noise~$\times$~R2L run that achieves a \emph{worse} minimum than R2L alone (3.910), suggesting that at 15\%, masking sufficiently disrupts the reversed sequence to negate permutation benefit.

\input{tables/tab_combinations}

\begin{figure}[t]
  \centering
  \includegraphics[width=\columnwidth]{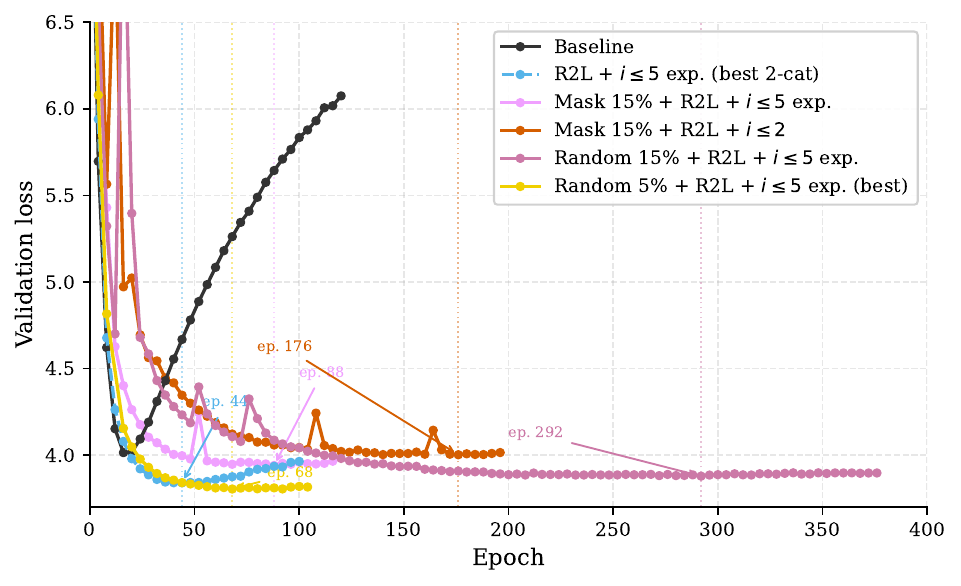}
    \vspace{-10pt}
  \caption{All 3-category combinations (solid) vs.\ the best 2-cat base R2L~+~$i{\leq}5$ exp.\ (dashed). Reducing the noise rate from 15\% to 5\% (gold) resolves the interference: Rand.\ 5\%~+~R2L~+~$i{\leq}5$ exp.\ achieves the overall best minimum of 3.805 at epoch~68, surpassing all individual and 2-category methods.}
    \vspace{-10pt}
  \label{fig:3cat}
\end{figure}

\paragraph{Three-category combinations: noise rate determines interference or synergy.}
Figure~\ref{fig:3cat} and the lower section of Table~\ref{tab:combinations} show all four 3-category combinations alongside the best 2-cat base, R2L~+~$i{\leq}5$ exp. When token noise is set at 15\%, adding it to any R2L~+~offset pair raises minimum loss while delaying the point of overfitting, the same trade-off observed in token noise~$\times$~permutation, now compounded. The degree of extension and cost depend heavily on the noise type and offset difficulty. Starting from R2L~+~$i{\leq}5$ exp.\ (epoch~44), adding mask~15\% delays overfitting to epoch~88 at a cost of 0.100 (3.841~$\to$~3.941). Adding random~15\% instead raises minimum loss by only 0.038 (3.841~$\to$~3.879), a much smaller penalty than masking. Starting from the cheaper R2L~+~$i{\leq}2$ base (epoch~96), adding mask~15\% delays overfitting to epoch~176 at a cost of 0.140 (3.863~$\to$~4.003). The random~15\%~+~R2L~50\%~+~$i{\leq}5$ exp.\ configuration also shows the most pronounced training instability of all runs, with recurring evaluation loss spikes throughout the stable phase. The three-way interaction of token corruption, sequence reversal, and offset prediction produces a highly complex and variable gradient signal.

However, we note that the token noise~$\times$~R2L analysis above already established that \emph{lower noise gives better minimum loss} in 2-category combinations. We therefore reduce the noise from 15\% to 5\%. This caused random~5\%~+~R2L~50\%~+~$i{\leq}5$ exp.\ to achieve a minimum loss of {3.805} at epoch~68, improving on the best individual method (random~15\%, 3.841) by 0.036 and beating all previously tested configurations. This suggests that at 5\%, token noise is mild enough that it no longer meaningfully corrupts the local context that offset prediction relies on; instead, the three objectives complement each other, each regularizing a different aspect of the training signal.

\begin{insightbox}
\textbf{Finding 5.} The largest gains come from composing augmentations, and compatibility decides the outcome: views that preserve each other's signal reinforce, while those that corrupt it interfere. Choosing augmentations that coexist matters more than choosing the strongest ones individually.
\end{insightbox}

\subsection{Do the rankings hold after decay?}
\label{sec:decay}

All prior comparisons used stable-phase checkpoints, which are evaluated before any learning-rate decay and may therefore not represent each configuration's true minimum. To verify that stable-phase rankings are meaningful proxies for fully-converged performance, we apply the WSD decay phase to all eight selected configurations. Full implementation details are provided in Appendix~\ref{sec:decay_details}. Figure~\ref{fig:decay} shows the resulting trajectories: solid lines are the stable-phase curves up to the decay start; dashed lines are the decay continuations; stars mark each decay minimum.

\begin{figure}[t]
  \centering
  \includegraphics[width=\columnwidth]{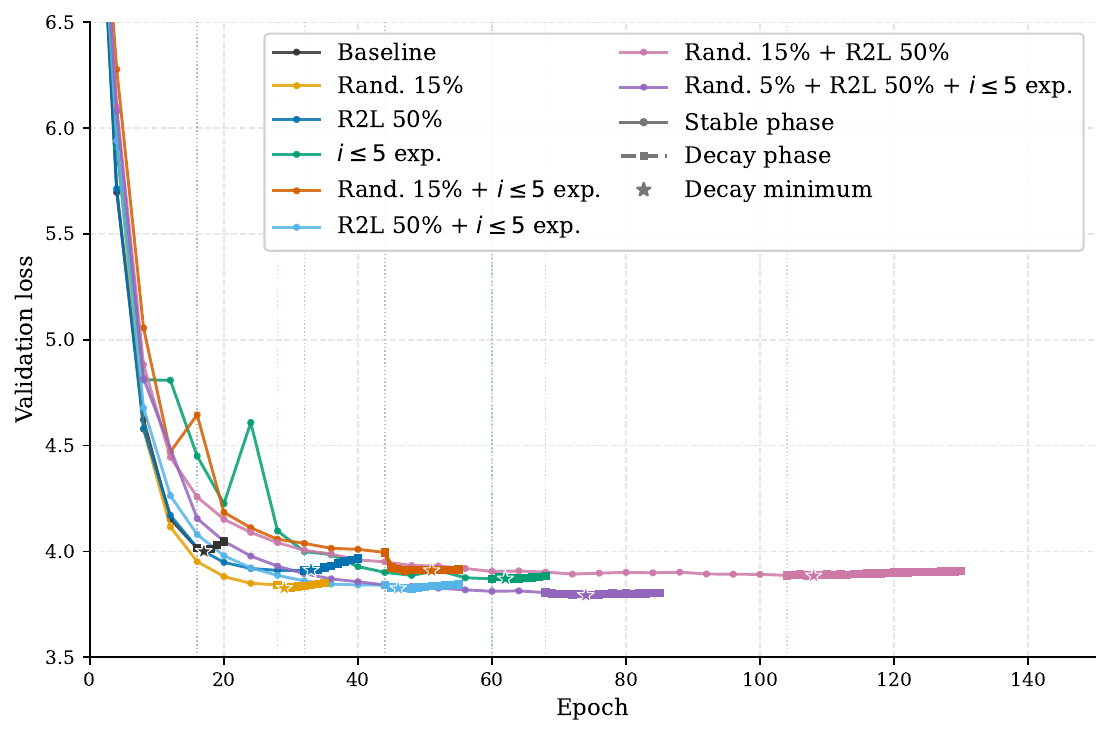}
  \caption{Validation loss trajectories for all eight configurations (epoch axis truncated at 150). Solid lines show the stable phase up to the WSD start; dashed lines show the decay continuation; stars mark each decay minimum. The 3-category combination (purple, Rand.\ 5\%+R2L+$i{\leq}5$ exp.) decays from epoch~68 and is fully visible; it reaches the lowest decay minimum of 3.792, the best result overall.}
  \label{fig:decay}
\end{figure}

Table~\ref{tab:decay} reports stable-phase and decay-phase minimum losses for all eight configurations. Several findings are notable as follows.

\input{tables/tab_decay}

\noindent \textbf{Rankings are largely preserved.} The stable-phase rank ordering is mostly reproduced after decay: Rand.\ 5\%~+~R2L~+~$i{\leq}5$~exp.\ is the clear best at both stable (3.805) and decay (3.792) phases, and the baseline stays last. The one notable reordering is that decay resolves the noise-offset interference in Rand.\ 15\%~+~$i{\leq}5$~exp.\ (3.995~$\to$~3.909), moving it ahead of R2L~50\% (3.912). This confirms that stable-phase comparisons are valid proxies for fully-converged performance, justifying their use as a cost-efficient experimental protocol throughout Sections~\ref{sec:individual_token}--\ref{sec:combinations}.

\input{tables/tab_zeroshot}

\noindent \textbf{Decay improvements are consistent with stable-phase results.} Six of the eight configurations improve by 0.017 or less after decay, indicating that stable-phase minima are already near the true optima. The 3-category combination (Rand.\ 5\%+R2L+$i{\leq}5$ exp.) achieves the best decay minimum of 3.792, improving its stable minimum by 0.013. The largest absolute improvement belongs to Rand.\ 15\%~+~$i{\leq}5$~exp.\ ($-0.086$), whose stable-phase minimum of 3.995 was severely inflated by interference between noise and offset prediction objectives; the LR decay resolves this conflict and allows the model to converge to 3.909. This confirms that the decay phase is particularly beneficial for configurations suffering from training-time objective interference.

\begin{insightbox}
\textbf{Finding 6.} Cheap stable-phase comparisons are a reliable proxy for fully-converged performance: method rankings are largely preserved through the final learning-rate decay, so augmentations can be compared without paying for the full decay each time.
\end{insightbox}

\subsection{Does lower loss improve downstream accuracy?}
\label{sec:downstream}

We evaluate eight fully-converged models on five zero-shot benchmarks: HellaSwag, PIQA, ARC-Challenge, WinoGrande, and COPA. Each model is evaluated at its decay-phase best checkpoint, i.e., the checkpoint with the lowest validation loss reached during the WSD decay (see Section~\ref{sec:decay}). For R2L~50\%, which saw no improvement under decay, we use the stable-phase minimum checkpoint instead.

\noindent \textbf{Benchmark selection.} We initially evaluated on ten tasks via \texttt{lm-evaluation-harness} but excluded five that produced no discriminative signal at 150M parameters; full details and justification are given in Appendix~\ref{sec:benchmark_details}. The five retained tasks are those that show meaningful variation across configurations.

\noindent \textbf{All augmented models outperform the baseline.} Every augmented configuration exceeds the baseline mean accuracy of 41.0\%, confirming that lower validation loss translates to improved generalization in aggregate. The margin ranges from modest (0.2\%) to significant (2.3\%), consistent with the limited model scale, but the direction is consistent across all seven augmented models.

\noindent \textbf{Validation loss rank does not perfectly predict accuracy rank.} The best val-loss model, Rand.\ 5\%~+~R2L~+~$i{\leq}5$~exp.\ (3.792), ranks second in mean accuracy (42.9\%). The top accuracy model, $i{\leq}5$~exp.\ (43.3\%), has the third-best val loss (3.870). This weak ordinal correlation is expected: at 150M parameters, individual task scores are highly variable, and differences of 0.05 in validation loss correspond to differences of only 1--2 pp in mean accuracy. We treat validation loss as the primary metric throughout this paper and downstream accuracy as corroborating evidence rather than a primary signal.

\noindent \textbf{Combination models show mixed downstream results.} Among 2-category combinations, only R2L~+~$i{\leq}5$~exp.\ achieves a competitive mean accuracy (42.1\%); the other two (Random~15\%~+~R2L and Random~15\%~+~$i{\leq}5$~exp.) reach only 41.2\%, barely above the baseline. This mirrors the val-loss picture: R2L~+~$i{\leq}5$~exp.\ is the only 2-cat combination that matched the best individual val loss (3.824), while the other two combinations suffered from interference. The 3-category combination (Rand.\ 5\%+R2L+$i{\leq}5$ exp.) performs well (42.9\%, second overall), with the highest scores on ARC-Challenge (27.6\%) and the second-highest on COPA (58.0\%) and HellaSwag (26.1\%). This suggests that training across a more diverse set of objectives improves generalization across reasoning domains. Overall, the downstream pattern reinforces the finding that combination hyperparameters matter: the Rand.\ 5\% variant achieves the best validation loss (3.792) and competitive downstream accuracy, while the Rand.\ 15\% variant sacrifices minimum loss for a longer stability plateau.

\begin{insightbox}
\textbf{Finding 7.} Lower validation loss translates into better downstream generalization in aggregate, so it is a sound optimization target; at this scale individual benchmark scores are too noisy to rank methods finely, and we read them only as corroboration.
\end{insightbox}

\section{Related Work}
\label{sec:related}

\paragraph{Data-constrained pretraining.}
As the supply of high-quality text tightens \cite{villalobos2022will, muennighoff2023scaling}, a growing body of work characterizes the multi-epoch regime and its failure modes: repeated data yields sharply diminishing returns \cite{muennighoff2023scaling}, drives memorization and degradation \cite{xue2023repeat, hernandez2022scaling, tirumala2022memorization}, and overfits even under abundant compute \cite{kim2025pre}. The proposed remedies act mostly at the data or architecture level, through data mixture and filtering \cite{su2025nemotron, soldaini2024dolma}, tuned dropout or Mixture-of-Experts \cite{xue2023repeat}, and regularization with ensembling \cite{kim2025pre}. We instead intervene on the training objective, varying the predictive view of each sequence.

\paragraph{Diffusion language models.}
Diffusion LMs resist multi-epoch overfitting far better than AR models \cite{prabhudesai2026diffusion, ni2025diffusion}, plausibly because denoising under varied corruption levels and factorization orders regularizes training. Adopting them wholesale, however, means leaving the autoregressive stack that current training and inference infrastructure is built around \cite{li2025survey, peng2025efficient, feng2026theoretical}. We keep the AR model and its left-to-right decoding and import only the regularizing mechanism, many stochastic views of the same data, through training-time augmentation.

\paragraph{Training-time objectives and augmentation.}
Many objectives already generate alternative views of a sequence: masked language modeling \cite{devlin2019bert}, permuted factorization orders \cite{yang2019xlnet}, the Mixture-of-Denoisers of UL2 \cite{tay2022ul2}, Fill-in-the-Middle \cite{bavarian2022fim}, and Meet-in-the-Middle \cite{nguyen2023meet}; instance-level data augmentation plays the same regularizing role in vision \cite{zhang2017mixup, cubuk2020randaugment}. These were developed for single-epoch, compute-constrained training, however, and were not studied as regularizers for high-epoch, fixed-corpus AR pretraining. The closest comparisons are the brief token-noise ablations in the diffusion-LM papers \cite{prabhudesai2026diffusion, ni2025diffusion}, which cover only one or two augmentations and are not their focus. Our contribution is a systematic study of which diffusion-style augmentations regularize AR pretraining in the data-constrained regime, with the architecture, loss, and left-to-right inference left unchanged. A fuller discussion of each direction appears in Appendix~\ref{sec:related_ext}.

\section{Conclusion and Discussion}
\label{sec:discussion}

Standard AR pretraining is fundamentally data-inefficient in the data-constrained regime: our baseline model reaches its minimum validation loss at epoch~16 and degrades continuously thereafter, making over 80\% of the available training budget counterproductive. This is a significant practical problem as the industry approaches a data ceiling. Our results demonstrate that data augmentation directly addresses this inefficiency. By serving as regularizers, augmentations allow the model to continue extracting useful signal from repeated passes over the same corpus: the best combination lowered the minimum validation loss from 4.015 to 3.805, a reduction of 0.210. These gains also translate to measurable downstream accuracy improvements across all augmented configurations. Despite this potential, data augmentation for LLM pretraining remains largely underexplored. Prior work has focused on data mixture and filtering strategies \cite{su2025nemotron, soldaini2024dolma, mohri2026bitter}, or adopted diffusion-style objectives as a wholesale replacement for AR training \cite{prabhudesai2026diffusion, ni2025diffusion}, rather than investigating augmentation as a lightweight regularizer within the standard AR framework. We hope this work helps establish the case that augmentation deserves serious attention as a first-class technique for pretraining in data-constrained settings.

Our experiments yield several concrete findings that can guide future work and practical application. For token-level noise, random token replacement consistently outperforms masking, a counterintuitive result given that masking is the conventional augmentation inspired by BERT-style models, which we attribute to (i) the increased difficulty of disambiguating a plausible-but-wrong token from context versus the explicit absence signal of a mask and (ii) the greater distribution shift between training and inference of the mask token. For sequence permutations, right-to-left prediction proves a strong regularizer while Fill-in-the-Middle provides no benefit in our general-domain setting, suggesting that augmentations which radically alter the sequence format relative to the evaluation distribution are less effective regularizers than those that preserve content and only reorder it. For combinations, our results reveal that orthogonal data augmentation methods can synergize to yield lower minimum losses than the baseline and any individual method. Nonetheless, we note that hyperparameter choices, particularly the noise rate, are decisive, and harder augmentations can cause evaluation loss spikes that may need attention.

\newpage
\bibliography{custom}

\newpage
\appendix

\section{Limitations}
\label{sec:limitations}

All experiments are conducted at a single model and data scale, i.e., a 150M-parameter Llama-based model trained on 75M tokens ($40\times$ below Chinchilla-optimal), due to compute constraints. It remains open whether the relative rankings among augmentation strategies generalize to larger models or to regimes closer to the Chinchilla-optimal data budget. The hyperparameter and combination coverage is also not exhaustive: only a subset of all possible 2- and 3-category combinations are explored, and finer-grained joint tuning may yield meaningfully better configurations. Finally, downstream evaluation is limited by model scale: at 150M parameters, zero-shot benchmark scores are highly variable, and differences of 0.05--0.10 in validation loss do not translate to reliable performance differences on individual tasks. Downstream accuracy should be interpreted as corroborating evidence rather than the primary signal.

\section{Future Work}
\label{sec:future_work}

The most immediate priority is scaling: repeating the ablations at multiple model sizes and data-to-parameter ratios would clarify whether the observed rankings are universal or regime-specific, and whether the interference patterns (e.g., token noise disrupting offset prediction) persist at scale. A complementary direction is dataset sensitivity: our experiments use a single web-text corpus without a model-based quality filter; testing on datasets with different domain distributions (e.g., C4, OSCAR) or with DCLM-style model-based filtering would establish how much the results depend on corpus composition. On the augmentation design side, a promising avenue is dynamic scheduling: rather than applying a fixed augmentation rate throughout training, one could ramp up augmentation intensity when the early-epoch memorization transition is detected, and reduce or disable augmentations during the decay phase to avoid conflicting gradient signals at convergence. Finally, it would be natural to combine training-time data augmentations with architecture-level regularizers such as dropout, which operates orthogonally at the activation level; understanding whether these two families of regularizers interact synergistically or redundantly would help practitioners build more effective multi-epoch training pipelines.

\section{Extended Related Work}
\label{sec:related_ext}

This section is an extended version of the Related Work in Section~\ref{sec:related}, giving a fuller account of each direction and how it relates to our study.

\paragraph{Autoregressive language model pretraining.}
The dominant paradigm for large language model pretraining is next-token prediction with a causal (left-to-right) autoregressive objective, established by the GPT line of models \cite{radford2019gpt2, brown2020gpt3} and carried forward by virtually all modern LLMs \cite{touvron2023llama, team2023gemini}. A central empirical finding has been that model quality scales predictably with both parameter count and dataset size \cite{kaplan2020scaling, hoffmann2022chinchilla}, motivating a sustained effort to curate ever-larger pretraining corpora \cite{li2024datacomp, soldaini2024dolma, su2025nemotron} and to generate synthetic training data at scale \cite{wang2025deeppersona}. Once pretrained, these models are commonly adapted to downstream tasks with parameter-efficient methods \cite{wang2023multitask}.

\paragraph{The data wall and AR inefficiency.}
This scaling strategy is approaching a hard limit. Analyses of the stock of high-quality public internet text project exhaustion within a few years at current consumption rates \cite{villalobos2022will, muennighoff2023scaling}, while GPU compute continues to grow faster than data availability. Compounding this problem is the inherent data inefficiency of standard AR pretraining. In response, a growing body of work studies training dynamics in data-constrained settings and proposes potential solutions to its unique problems. \citet{muennighoff2023scaling} show that repeated passes over the same corpus yield diminishing and eventually negligible returns beyond roughly four epochs, and then derive scaling laws for epoched training and recommend data mixture strategies. \citet{kim2025pre} demonstrate that standard data-repetition recipes eventually suffer from overfitting in a data-constrained, infinite-compute regime. To mitigate this, they introduce a framework utilizing heavily tuned regularization and ensemble scaling to drastically improve data efficiency, proving that these scaling gains can be successfully distilled into much smaller student models. \citet{xue2023repeat} similarly find that repeatedly training on the same data leads to severe overfitting and performance degradation. They show that carefully tuned dropout can effectively mitigate multi-epoch degradation, and recommend leveraging Mixture-of-Experts (MoE) models, part of a broader trend toward compositional models assembled from specialized experts \cite{kang2025selfmoe}.

\paragraph{Diffusion language models as an alternative.}
One response to the data-constrained challenge has been to abandon AR training in favour of diffusion language models. \citet{prabhudesai2026diffusion} and \citet{ni2025diffusion} independently demonstrate that diffusion LMs are substantially more robust to overfitting than AR models in high-epoch settings, and hypothesize that requiring the model to denoise under arbitrary corruption levels and factorization orders acts as a natural regularizer. However, diffusion models face significant practical barriers: fixed-length generation, the absence of KV-cache support, and the lack of mature inference infrastructure make large-scale deployment difficult \cite{li2025survey, peng2025efficient, feng2026theoretical}. Our work pursues a complementary direction: rather than replacing AR training, we ask whether diffusion-inspired augmentations can be imported into the AR framework as regularizers, preserving all existing infrastructure.

\paragraph{Training-time augmentation objectives.}
Several prior works have explored non-standard prediction objectives at pretraining time. BERT \cite{devlin2019bert} introduced masked language modeling, training a bidirectional encoder to recover corrupted tokens, but targets the fine-tuning regime and does not evaluate multi-epoch dynamics. XLNet \cite{yang2019xlnet} generalizes AR pretraining to arbitrary factorization orders via permutation language modeling. UL2 \cite{tay2022ul2} proposes a Mixture-of-Denoisers (MoD) framework that unifies span corruption, causal LM, and prefix LM objectives under a single model. Fill-in-the-Middle \cite{bavarian2022fim} introduces PSM/SPM sequence reordering as a code pretraining technique. \citet{nguyen2023meet} propose ``Meet in the Middle'' (MIM) pretraining, which jointly trains a L2R and a R2L model on the same data and encourages them to agree on their token predictions at each position, improving data efficiency and infilling capability. Despite these contributions, none of the above has received widespread adoption in general-purpose LLM pretraining pipelines. A key reason is that these works were developed in the \textit{compute-constrained} regime: their goal was to reduce validation loss within a fixed single-epoch budget, not to regularize across tens or hundreds of epochs. The setting and motivation therefore differ fundamentally from ours. The closest overlap with our work is in the data-constrained ablations performed by \citet{prabhudesai2026diffusion} and \citet{ni2025diffusion}, who include brief comparisons of token-noise augmentations applied to AR baselines as foils for their diffusion models. These ablations are limited to one or two augmentation types, are not systematic across hyperparameter choices, and are not the focus of either paper, leaving a significant gap for practitioners wishing to understand how to apply data augmentation to AR pretraining. Other work analyzes the prediction behavior of trained AR models, for example showing that they can underweight long-range context and correcting this at inference time \cite{malkin2022coherence}.

\paragraph{Data augmentation in computer vision.}
Data augmentation has been foundational to the success of deep learning in computer vision for over a decade \cite{shorten2019survey}. Techniques such as random cropping, flipping, color jitter, CutMix \cite{yun2019cutmix}, MixUp \cite{zhang2017mixup}, and RandAugment \cite{cubuk2020randaugment} are standard components of state-of-the-art image classifiers and self-supervised vision models \cite{chen2020simple, caron2021emerging}. The underlying mechanism is directly analogous to our setting: augmentations increase the effective diversity of the training distribution, acting as regularizers that reduce overfitting when the model has more capacity or training compute than the raw dataset can fully utilize. The success of augmentations in vision suggests substantial untapped potential in the language domain, where training-time instance-level transformations have received far less systematic study.

\section{Training Details}
\label{sec:training_details}

Our model is a decoder-only causal language model following the Llama architecture \cite{touvron2023llama}. The architecture uses pre-normalization with RMSNorm ($\epsilon = 10^{-6}$), SwiGLU feed-forward blocks \cite{shazeer2020glu}, and rotary positional embeddings (RoPE) \cite{su2024roformer}. We use tied input/output embeddings to keep the parameter count manageable given the large Qwen2 vocabulary.

The model hyperparameters are summarized in Table~\ref{tab:arch}. They are inspired by and extrapolated from the DCLM scaling recipe, which specifies a consistent head dimension of $d_{\text{head}} = 128$ and an intermediate size following the SwiGLU formula $d_{\text{ffn}} = \frac{8}{3} d_{\text{model}}$ rounded to the nearest multiple of 256. At 150M parameters our model falls below the smallest DCLM competition scale (412M), so our hyperparameters are linearly extrapolated downward: we reduce the number of layers to 20 and the model width to 512 while preserving the same $d_{\text{head}}=128$ ratio (giving 4 attention heads), and set the intermediate size to 1,536 ($\approx 3\times512$, consistent with the DCLM formula).

\input{tables/tab_arch}

We use the AdamW optimizer \cite{loshchilov2017decoupled} with the hyperparameters listed in Table~\ref{tab:optim}. We use a peak learning rate of $6\times10^{-4}$, a weight decay of 0.033, and 100 warmup steps. Gradient norm clipping is applied at 1.0. The global batch size is 512 sequences of 2,048 tokens.

\input{tables/tab_optim}

We use the Qwen2 tokenizer (vocabulary size 151,646) \cite{yang2025qwen3}. The large vocabulary is one motivation for tied embeddings: at 512 hidden size, the embedding matrix alone accounts for $151{,}646 \times 512 \approx 77$M parameters, so tying input and output embeddings halves this contribution and keeps the total parameter count near 150M. Augmentation-specific special tokens are added to the vocabulary as needed: a direction pair (\texttt{<|l2r\_pred|>}/\texttt{<|r2l\_pred|>}), per-offset tokens (\texttt{<|next\_i\_pred|>} for each $i \leq n$), a mask token (\texttt{<|mask|>}), and three FIM control tokens (\texttt{<|fim\_prefix|>}, \texttt{<|fim\_suffix|>}, \texttt{<|fim\_middle|>}). These tokens are protected from corruption by the token-noise augmentation.

Training is implemented using the HuggingFace \texttt{Trainer} \cite{wolf2020transformers} on either a 4xA100 or 2xH100 GPU setup. Evaluation checkpoints are saved every 4 epochs during stable-phase training and at every epoch during decay-phase.

\section{Held-Out Validation Details}
\label{sec:val_details}

Our primary metric is held-out validation loss computed on a fixed validation dataset, acquired from a different shard of DCLM-RefinedWeb \cite{li2024datacomp} than the training split. Both splits use the same preprocessing pipeline: documents are tokenized and packed into contiguous 2,048-token blocks with remainder tokens discarded, so individual examples may span document boundaries. Validation is always evaluated under standard left-to-right next-token prediction ($i=1$) with all training-time augmentations disabled; direction and offset control tokens are prepended only when the model was trained with them. We did not apply additional near-duplicate filtering between the training and validation splits, considering there are zero URL overlap and only 45 exact-text duplicates ($\approx$0.8\% of validation documents, mostly boilerplate web pages with different URLs). Corpus-level deduplication and quality filtering are inherited from the DCLM dataset pipeline.

\section{Decay-Phase Training Details}
\label{sec:decay_details}

\textbf{Checkpoint selection.} For each of the eight configurations, we scan all evaluation checkpoints recorded during the stable training phase and identify the checkpoint with the lowest held-out validation loss. This checkpoint serves as the resume point for the WSD decay. Table~\ref{tab:decay_details} lists the resume epoch and the corresponding global training step for each configuration.

\textbf{Decay schedule.} Starting from the resume checkpoint, we apply the $1{-}\sqrt{\cdot}$ learning-rate schedule as recommended by \citet{hagele2024scaling}. Letting $n$ denote the current training step, $N$ the last training step, and $N_{\text{decay}}$ the number of decay steps, the schedule multiplies the peak rate by
\begin{equation}
f(n,\,N,\,N_{\text{decay}}) = 1 - \sqrt{\frac{n - (N - N_{\text{decay}})}{N_{\text{decay}}}},
\label{eq:decay}
\end{equation}
decaying from 1 at the start of the decay phase ($n = N - N_{\text{decay}}$) to 0 at the final step ($n = N$). We set $N_{\text{decay}} \approx 20\%$ of the stable-phase training steps; the exact values per run are listed in Table~\ref{tab:decay_details}. All other hyperparameters (batch size, weight decay, AdamW $\beta$ values, context length) are kept identical to the stable phase, and no additional warmup is applied at the resume point. The final converged loss reported in Table~\ref{tab:decay} is the minimum validation loss observed at any checkpoint during the decay run.

\textbf{Resume checkpoints.} Table~\ref{tab:decay_details} lists the resume checkpoint (global step and corresponding epoch) and the total number of decay steps for each configuration.

\input{tables/tab_decay_details}

\section{Downstream Evaluation Details}
\label{sec:benchmark_details}

We evaluate using \texttt{lm-evaluation-harness} \cite{eval-harness} in zero-shot mode. All tasks use the model's standard left-to-right next-token prediction setting; no task-specific fine-tuning or prompting is applied. Table~\ref{tab:benchmarks} lists the five benchmarks retained for the final analysis, with their category, description, and source.

\input{tables/tab_benchmarks}

\textbf{Exclusion rationale.} At 150M parameters, several standard benchmarks saturate or collapse in ways that yield no discriminative signal. LAMBADA requires resolving long-range dependencies that are beyond the reach of a small model, producing 0\% accuracy universally. BoolQ, RTE, and CommonsenseQA all collapse to the majority-class prediction regardless of augmentation, indicating the model has not learned task-relevant representations. ARC-Easy and OpenBookQA both show near-identical scores across all eight configurations (variance $<$0.5~pp), providing no basis for comparison. SciQ sits near-random ($\approx$5\%), likely because it requires passage retrieval that zero-shot evaluation cannot support. The five retained tasks (HellaSwag, PIQA, ARC-Challenge, WinoGrande, COPA) all show at least 2~pp of variation across configurations and a plausible ordering relative to validation loss, making them the most informative signal available at this scale.

\end{document}

%% file: tables/tab_individual.tex
\begin{table}[t]
\centering
\small
\setlength{\tabcolsep}{4pt}
\begin{tabular}{lcc}
\hline
\textbf{Method} & \textbf{Min loss} & \textbf{Best ep.} \\
\hline
Baseline & 4.015 & \phantom{0}16 \\
\hline
\multicolumn{3}{l}{\textit{Token-level noise}} \\
Mask 15\%          & 3.910 & \phantom{0}28 \\
Mask 30\%          & 3.923 & \phantom{0}40 \\
Random 5\%         & 3.912 & \phantom{0}20 \\
Random 15\%        & \textbf{3.841} & \phantom{0}28 \\
Random 30\%        & 3.865 & \phantom{0}44 \\
\hline
\multicolumn{3}{l}{\textit{Sequence permutation}} \\
R2L 50\%           & 3.910 & \phantom{0}32 \\
R2L 25\%           & 3.942 & \phantom{0}24 \\
FIM 50\%           & 3.947 & \phantom{0}16 \\
\hline
\multicolumn{3}{l}{\textit{Target offset prediction}} \\
$i{\leq}2$, uniform & 3.890 & \phantom{0}24 \\
$i{\leq}5$, uniform & 4.016 & \phantom{0}32 \\
$i{\leq}5$, exp.    & 3.870 & \phantom{0}60 \\
\hline
\end{tabular}
\caption{Minimum stable-phase validation loss and corresponding epoch for all individual augmentation methods. \textbf{Bold} = best individual result.}
\label{tab:individual}
  \vspace{-10pt}
\end{table}

%% file: tables/tab_combinations.tex
\begin{table}[t]
\centering
\small
\setlength{\tabcolsep}{4pt}
\begin{tabular}{lcc}
\hline
\textbf{Combination} & \textbf{Min loss} & \textbf{Best ep.} \\
\hline
\multicolumn{3}{l}{\textit{Permutation $\times$ offset (synergy)}} \\
R2L 50\% + $i{\leq}5$ exp. & 3.841 & \phantom{0}44 \\
R2L 50\% + $i{\leq}2$      & 3.863 & \phantom{0}96 \\
\hline
\multicolumn{3}{l}{\textit{Token noise $\times$ permutation}} \\
Random 5\% + R2L 50\%  & 3.877 & \phantom{0}48 \\
Random 15\% + R2L 50\% & 3.887 & 104 \\
Mask 5\% + R2L 50\%    & 3.897 & \phantom{0}40 \\
Mask 15\% + R2L 50\%   & 3.963 & \phantom{0}64 \\
\hline
\multicolumn{3}{l}{\textit{Token noise $\times$ offset (interference)}} \\
Mask 5\% + $i{\leq}2$         & 3.937 & \phantom{0}24 \\
Random 15\% + $i{\leq}5$ exp. & 3.995 & \phantom{0}44 \\
Mask 15\% + $i{\leq}2$        & 4.004 & \phantom{0}40 \\
\hline
\multicolumn{3}{l}{\textit{3-category combinations}} \\
Rand.\ 5\%+R2L+$i{\leq}5$ exp.  & \textbf{3.805} & \phantom{0}68 \\
Rand.\ 15\%+R2L+$i{\leq}5$ exp. & 3.879 & 292 \\
Mask 15\%+R2L+$i{\leq}5$ exp.   & 3.941 & \phantom{0}88 \\
Mask 15\%+R2L+$i{\leq}2$        & 4.003 & 176 \\
\hline
\end{tabular}
\caption{All 2- and 3-category combination results grouped by interaction type. See Table~\ref{tab:individual} for individual method baselines. \textbf{Bold} = best combination result; note that Rand.\ 5\%+R2L+$i{\leq}5$ exp.\ beats all individual ones.}
\label{tab:combinations}
  \vspace{-10pt}
\end{table}

%% file: tables/tab_decay.tex
\begin{table}[t]
  \centering
  \footnotesize
  \setlength{\tabcolsep}{3pt}
  \begin{tabular}{lccc}
  \hline
  \textbf{Run} & \textbf{Stable} & \textbf{Decay} & $\boldsymbol{\Delta}$ \\
  \hline
  Baseline                      & 4.015  & 4.000  & $-$0.015 \\
  \hline
  Rand.\ 15\%                   & 3.841  & 3.826  & $-$0.015 \\
  R2L 50\%                      & 3.910  & 3.912  & $+$0.002 \\
  $i{\leq}5$ exp.               & 3.870  & 3.870  & \phantom{$-$}0.000 \\
  \hline
  Rand.\ 15\%+R2L               & 3.887 & 3.884  & $-$0.003 \\
  Rand.\ 15\%+$i{\leq}5$ exp.   & 3.995  & 3.909  & $\mathbf{-0.086}$ \\
  R2L+$i{\leq}5$ exp.           & 3.841  & 3.824  & $-$0.017 \\
  \hline
  Rand.\ 5\%+R2L+$i{\leq}5$ exp. & \textbf{3.805} & \textbf{3.792} & $-$0.013 \\
  \hline
  \end{tabular}
  \caption{Stable-phase and decay-phase minimum validation loss, and improvement $\Delta$. Negative $\Delta$ = decay lowers loss. \textbf{Bold} = best per column.}
  \label{tab:decay}
    \vspace{-10pt}
\end{table}

%% file: tables/tab_zeroshot.tex
\begin{table*}[t]
\centering
\small
\setlength{\tabcolsep}{6pt}
\begin{tabular}{lcccccccc}
\hline
\textbf{Run} & \textbf{Val loss} & \textbf{HellaSwag} & \textbf{PIQA} & \textbf{ARC-C} & \textbf{WinoGrande} & \textbf{COPA} & \textbf{Mean} \\
\hline
Baseline                              & 4.000 & 25.1 & \underline{56.0} & 24.9 & 48.7 & 50.0 & 41.0 \\
\hline
Random repl.\ 15\%                    & 3.826 & 25.5 & \textbf{57.6} & 25.4 & 50.1 & 54.0 & 42.5 \\
R2L 50\%                              & 3.910 & 25.6 & 54.0 & 24.8 & \textbf{52.0} & 56.0 & 42.5 \\
$i{\leq}5$ exp.                       & 3.870 & 25.4 & 53.8 & 25.4 & \textbf{52.0} & \textbf{60.0} & \textbf{43.3} \\
\hline
Random 15\% + R2L 50\%                & 3.884 & \textbf{26.2} & 53.2 & 25.3 & \underline{51.5} & 50.0 & 41.2 \\
Random 15\% + $i{\leq}5$ exp.        & 3.909 & 26.0 & 52.8 & 24.2 & 50.7 & 52.0 & 41.2 \\
R2L 50\% + $i{\leq}5$ exp.           & \underline{3.824} & 25.9 & 52.3 & \underline{27.5} & 50.0 & 55.0 & 42.1 \\
\hline
Rand.\ 5\%+R2L 50\%+$i{\leq}5$ exp.  & \textbf{3.792} & \underline{26.1} & 53.2 & \textbf{27.6} & 49.6 & \underline{58.0} & \underline{42.9} \\
\hline
\end{tabular}
\caption{Post-decay validation loss and zero-shot accuracy (\%) on five benchmarks for all eight configurations. Accuracy is normalized (\texttt{acc\_norm}) for HellaSwag, PIQA, ARC-Challenge, and WinoGrande; standard accuracy for COPA. \textbf{Bold} = best per column; \underline{underline} = second best.}
\label{tab:zeroshot}
  \vspace{-10pt}
\end{table*}

%% file: tables/tab_arch.tex
\begin{table}[h]
\centering
\small
\setlength{\tabcolsep}{5pt}
\begin{tabular}{lr}
\hline
\textbf{Hyperparameter} & \textbf{Value} \\
\hline
Parameters             & $\approx$150M \\
Layers ($n_{\text{layers}}$)          & 20 \\
Attention heads ($n_{\text{heads}}$)  & 4 \\
Model width ($d_{\text{model}}$)      & 512 \\
Head dimension ($d_{\text{head}}$)    & 128 \\
Intermediate size ($d_{\text{ffn}}$)  & 1,536 \\
Context length                        & 2,048 \\
Tokenizer                             & Qwen2 (vocab 151,646) \\
\hline
\end{tabular}
\caption{Model architecture hyperparameters. The head dimension and intermediate size follow the DCLM scaling recipe \cite{li2024datacomp}.}
\label{tab:arch}
\end{table}

%% file: tables/tab_optim.tex
\begin{table}[h]
\centering
\small
\setlength{\tabcolsep}{5pt}
\begin{tabular}{lr}
\hline
\textbf{Hyperparameter} & \textbf{Value} \\
\hline
Optimizer              & AdamW \\
$\beta_1$, $\beta_2$   & 0.9, 0.999 \\
$\epsilon$             & $10^{-8}$ \\
Peak learning rate     & $6 \times 10^{-4}$ \\
LR schedule (stable)   & Constant \\
LR schedule (decay)    & $1{-}\sqrt{\cdot}$ (Eq.~\ref{eq:decay}) \\
Warmup steps           & 100 (linear) \\
Weight decay           & 0.033 \\
Gradient clip norm     & 1.0 \\
Batch size & 512 \\
\hline
\end{tabular}
\caption{Optimization hyperparameters. LR and weight decay follow the DCLM recipe \cite{li2024datacomp}.}
\label{tab:optim}
\end{table}

%% file: tables/tab_decay_details.tex
\begin{table}[ht]
\centering
\footnotesize
\setlength{\tabcolsep}{3pt}
\begin{tabular}{lcc}
\hline
\textbf{Run} & \textbf{Resume step (ep.)} & \textbf{Decay steps} \\
\hline
Baseline                        & 1{,}152 (16)   & $\approx$230 \\
Rand.\ 15\%                     & 2{,}016 (28)   & $\approx$403 \\
R2L 50\%                        & 2{,}304 (32)   & $\approx$461 \\
$i{\leq}5$ exp.                 & 4{,}320 (60)   & $\approx$864 \\
Rand.\ 15\%+R2L                 & 7{,}488 (104)  & $\approx$1{,}498 \\
Rand.\ 15\%+$i{\leq}5$ exp.      & 3{,}168 (44)   & $\approx$634 \\
R2L+$i{\leq}5$ exp.              & 3{,}168 (44)   & $\approx$634 \\
Rand.\ 5\%+R2L+$i{\leq}5$ exp.   & 4{,}896 (68)   & $\approx$979 \\
\hline
\end{tabular}
\caption{Resume checkpoint (global step and epoch in parentheses) and $N_{\text{decay}}$ (Eq.~\ref{eq:decay}), set to approximately 20\% of the resume step, for each of the eight decay runs.}
\label{tab:decay_details}
\end{table}

%% file: tables/tab_benchmarks.tex
\begin{table*}[t]
\centering
\small
\setlength{\tabcolsep}{5pt}
\begin{tabular}{llp{5.5cm}l}
\hline
\textbf{Task} & \textbf{Category} & \textbf{Description} & \textbf{Source} \\
\hline
HellaSwag     & Commonsense NLI       & Choose the most plausible completion for a partial activity description (4-way MC). & \citet{zellers2019hellaswag} \\
PIQA          & Physical reasoning    & Choose the more physically plausible procedure for an everyday goal (2-way MC). & \citet{bisk2020piqa} \\
ARC-Challenge & Science QA            & Grade-school science multiple-choice questions, harder partition requiring reasoning beyond recall. & \citet{clark2018think} \\
WinoGrande    & Commonsense reasoning & Fill-in-the-blank pronoun coreference requiring commonsense knowledge (2-way MC). & \citet{sakaguchi2021winogrande} \\
COPA          & Causal reasoning      & Choose the most plausible cause or effect of a given premise (2-way MC). & \citet{roemmele2011choice} \\
\hline
\end{tabular}
\caption{Downstream benchmarks retained for evaluation via \texttt{lm-evaluation-harness}. All tasks are scored by accuracy; for HellaSwag, PIQA, ARC-Challenge, and WinoGrande we report length-normalised accuracy (\texttt{acc\_norm}), and for COPA we report standard accuracy (\texttt{acc}).}
\label{tab:benchmarks}
\end{table*}